\documentclass[11pt]{article}

\usepackage{amsmath,amssymb,amsfonts}
\usepackage{graphicx}
\usepackage{hyperref}
\usepackage{natbib}
\usepackage{color}
\usepackage{amsmath,amssymb}
\usepackage{amsthm}
\usepackage{algorithm}
\usepackage{algpseudocode}
\usepackage{caption}

\newtheorem{theorem}{Theorem}[section]

\newtheorem{proposition}[theorem]{Proposition}

\newtheorem{definition}[theorem]{Definition}

\usepackage[margin=1in]{geometry}

\title{NePPO: Near-Potential Policy Optimization for General-Sum Multi-Agent Reinforcement Learning}

\author{
Addison Kalanther \thanks{Electrical Engineering and Computer Sciences, UC Berkeley, \textbf{email:} addikala@berkeley.edu } \and
Sanika Bharvirkar \thanks{Electrical Engineering and Computer Sciences, UC Berkeley, \textbf{email:} sbharvirkar@berkeley.edu} \and 
Shankar Sastry \thanks{Electrical Engineering and Computer Sciences, UC Berkeley, \textbf{email:}  sastry@coe.berkeley.edu } \and 
Chinmay Maheshwari \thanks{Electrical and Computer Engineering, Data Science  \& Artificial Intelligence Institute, Johns Hopkins University \textbf{email:} chinmay\_maheshwari@jhu.edu}
\thanks{Chinmay Maheshwari would like to acknowledge the support of AWS Agentic AI Amazon Research Award 2026.}
}

\date{}

\begin{document}

\maketitle

\begin{abstract}
Multi-agent reinforcement learning (MARL) is increasingly used to design learning-enabled agents that interact in shared environments. However, training MARL algorithms in general-sum games remains challenging: learning dynamics can become unstable, and convergence guarantees typically hold only in restricted settings such as two-player zero-sum or fully cooperative games. Moreover, when agents have heterogeneous and potentially conflicting preferences, it is unclear what system-level objective should guide learning. In this paper, we propose a new MARL pipeline called \textbf{Ne}ar-\textbf{P}otential \textbf{P}olicy \textbf{O}ptimization (NePPO) for computing approximate Nash equilibria in mixed cooperative--competitive environments. The core idea is to learn a player-independent potential function such that the Nash equilibrium of a cooperative game with this potential as the common utility approximates a Nash equilibrium of the original game. To this end, we introduce a novel MARL objective such that minimizing this objective yields the best possible potential function candidate and consequently an approximate Nash equilibrium of the original game. We develop an algorithmic pipeline that minimizes this objective using zeroth-order gradient descent and returns an approximate Nash equilibrium policy. We empirically show the superior performance of this approach compared to popular baselines such as IPPO and MAPPO. \footnote{Our code is publicly available at \url{https://github.com/adkala/neppo}.}

\end{abstract}
\section{Introduction}
Multi-agent reinforcement learning (MARL) provides a principled framework for designing autonomous systems in which multiple learning-enabled agents interact and adapt in dynamic environments. Such settings arise in a wide range of domains, including autonomous racing \citep{kalaria2025alpharacer}, aerial pursuit \citep{kalantheradaptivePEG}, autonomous driving \citep{shalev2016safe}, and large-scale logistics systems \citep{krnjaic2024scalable}. In these environments, each agent's decision alters the effective environment faced by others, leading to strategically coupled training dynamics. Moreover, agents often possess heterogeneous—and potentially conflicting—objectives, which fundamentally distinguishes MARL from single-agent reinforcement learning.

A central solution concept for analyzing such interactions is the Nash equilibrium \citep{zhang2021multi,albrecht2024multi}, defined as a policy profile in which no agent can improve its utility via unilateral deviation. While Nash equilibria provide a natural notion of stability, existing MARL algorithms are provably guaranteed to converge to Nash equilibria only in restricted settings, such as two-player zero-sum games \citep{zhang2024survey} or fully cooperative games \citep{zhong2024heterogeneous}. In general-sum settings, learning dynamics can exhibit non-convergent or chaotic behavior \citep{MazumdarPolicyGradientNonConvergence,li2025multi}, and even when convergence occurs, equilibrium selection remains ambiguous due to the presence of multiple equilibria with potentially undesirable outcomes.
In this work, we ask the question: Can one design a tunable training objective that captures heterogeneous agent preferences while promoting convergence to (approximate) Nash equilibria?

In this paper, we address this challenge through the framework of \emph{Markov Near-Potential Functions (MNPFs)} \citep{guo2025markov,maheshwari2024convergence,guo2024alpha,guo2025distributed,guo2025bsde}. An MNPF is a player-independent function that approximates the change in each agent's value induced by unilateral deviations (see Definition \ref{def:near_pot_def}). A key property of this framework is that a maximizer of the potential function corresponds to an approximate Nash equilibrium of the original game (see Proposition \ref{prop:MNPF}).

Building on this idea, we introduce a novel MARL pipeline that learns a potential function and uses it to compute approximate Nash equilibria. Our key departure from prior formulations of Markov near-potential functions in \citep{guo2025markov,maheshwari2024convergence,guo2024alpha,guo2025distributed,guo2025bsde} is that we only require the potential function to approximate changes in players' utility due to unilateral deviation around a reference policy, rather than uniformly over the entire policy space. Here, for any potential function candidate \(\Phi\), the reference policy (denoted by \(\pi^{\ast,\Phi}\)) is defined as a Nash equilibrium of a cooperative game in which all agents optimize the potential function as shared objective.

Formally, we define a new MARL objective (see \eqref{eq:relaxed_F_i}) over candidate potential functions $\Phi$, which, for each agent $i$, measures the discrepancy between:
\begin{enumerate}
    \item the change in $\Phi$ induced by a unilateral best-response deviation from $\pi^{*,\Phi}$, and
    \item the corresponding change in the player's value function.
\end{enumerate}
We show that if, for some \(\Phi\), this discrepancy is bounded by $\alpha \geq 0$, for every player, then $\pi^{*,\Phi}$ constitutes an $\alpha$-approximate Nash equilibrium of the original game (Theorem~\ref{thm:ApproxNashThroughPotentialNew}). 
Moreover, the parameterization of the potential function $\Phi$ can serve as a {tunable design handle} to accommodate other system objective.\footnote{In particular, restricting $\Phi$ to a structured function class implicitly biases the set of attainable equilibria, thereby enabling the designer to encode system-level objectives directly through the representation of $\Phi$, without modifying individual agents' reward functions.} 

From an algorithmic standpoint, we develop a zeroth-order optimization procedure for minimizing the proposed objective (Algorithm~\ref{alg:nppg}). The resulting method relies on two modular computational components:
\begin{itemize}
    \item[(M1)] Solving a cooperative MARL problem in which all agents maximize the learned potential function;
    \item[(M2)] Computing approximate best responses for individual agents.
\end{itemize}
This modular structure allows us to leverage existing MARL algorithms. In particular, (M1) can be implemented using cooperative methods such as HAPPO \citep{zhong2024heterogeneous}, while (M2) can be instantiated using standard policy-gradient techniques such as PPO \citep{schulman2017proximal}.
Finally, we validate our approach through numerical experiments, demonstrating improved performance in terms of equilibrium quality and stability compared to standard MARL baselines.

\textbf{Prior Work.} 
Early advances in multi-agent reinforcement learning (MARL) emerged from classical board and card games such as Chess, Go, and Poker
\citep{campbell2002deep, schaeffer1992world, sheppard2002world, silver2016mastering}.
These discrete, fully observable settings enabled major algorithmic progress in search and counterfactual regret minimization (CFR) methods (e.g. \citep{zinkevich2007regret, tammelin2014solving, brown2016strategy}) by providing convergence guarantees to Nash equilibria in two-player zero-sum games.
However, these guarantees rely heavily on the special structure of zero-sum normal-form games. They do not extend to continuous-action, partially observable, or general-sum environments that often characterize real-world autonomous systems.

Multi-player video games have emerged as large-scale benchmarks for mixed cooperative-competitive MARL.
Population-based self-play in DeepMind’s Capture-the-Flag environment \citep{jaderberg2019human} demonstrated the emergence of coordinated behaviors under partial observability.
Subsequently, this paradigm was extended to complex competitive domains such as StarCraft~II (AlphaStar) \citep{vinyals2019grandmaster, wang2021scc} and Dota~2 (OpenAI Five) \citep{berner2019dota}.
Additional studies explored emergent cooperation and competition in physics-based environments such as Hide-and-Seek \citep{baker2019emergent}, continuous-control benchmarks built on MuJoCo and PettingZoo \citep{gupta2017cooperative, terry2020pettingzoo}, and team-sports simulations such as Google Research Football \citep{kurach2020google, lin2023tizero}.
While these environments enabled training at an unprecedented scale, self-play optimization in such systems often exhibits cycling, unstable co-adaptation, or brittle strategic behavior \citep{xu2023fictitious, zhang2024survey}. 

To address instability in self-play training, the \emph{Policy-Space Response Oracles} (PSRO) framework \citep{lanctot2017unified} introduced a principled population-based approach grounded in game theory.
PSRO iteratively generates approximate best responses and computes a meta-strategy over the resulting policies, approximating a Nash equilibrium of the empirical game.
Subsequent work developed scalable variants in \citep{mcaleer2020pipeline, mcaleer2022anytime}.
Despite these advances, most theoretical guarantees remain restricted to static normal-form games, leaving convergence and equilibrium selection in dynamic, continuous-action environments largely unresolved.

In parallel, the centralized-training, decentralized-execution (CTDE) paradigm adapted deep reinforcement learning methods to multi-agent systems 
Algorithms such as MADDPG, COMA, QMIX, and MAPPO stabilize learning through centralized critics or value-factorization while enabling decentralized policies at execution time.
While effective in cooperative or homogeneous-agent settings, these approaches typically lack principled equilibrium objectives or termination criteria in heterogeneous and mixed-motive environments \citep{terry2020revisiting, MazumdarPolicyGradientNonConvergence, li2025multi, zhong2024heterogeneous, hu2023marllib}.
Recent theoretically grounded variants such as HAML \citep{zhong2024heterogeneous} provide convergence guarantees primarily under cooperative assumptions.

\noindent
Despite significant progress in these algorithmic paradigms, the fundamental challenge of designing tunable training objective that captures heterogeneous agent preferences
remain unresolved.
Existing approaches either rely on restrictive game structures (e.g., zero-sum or cooperative settings) or focus on empirical performance without equilibrium guarantees.

\section{Background}
Consider a multi-agent environment comprising of $N$ agents interacting with each other under partially observable (heterogeneous) information. Such interactions are mathematically studied under the framework of Partially-Observable Markov Games (POMGs) \cite{albrecht2024multi}. We consider an infinite-horizon discounted POMG characterized by the tuple 
$
\mathcal{G} = \langle \mathcal{N}, \mathcal{S}, (\mathcal{A}_i)_{i\in \mathcal{N}}, \mathcal{P},$ $(\mathcal{Z}_i)_{i\in \mathcal{N}}, \mathcal{O}, (r_i)_{i\in\mathcal{N}}, \gamma, \rho \rangle,
$
where $\mathcal{N} = \{1,\dots,N\}$ denotes the set of agents; $\mathcal{S} \subseteq \mathbb{R}^{d_s}$ denotes the (continuous) state space; $\mathcal{A}_i \subseteq \mathbb{R}^{d_{a_i}}$ denotes the (continuous) action space of agent $i \in \mathcal{N}$; $\mathcal{A} = \times_{i\in \mathcal{N}} \mathcal{A}_i$ denotes the joint action space; $\Delta(\mathcal{S})$ denotes the set of probability measures on $\mathcal{S}$; $\mathcal{P}: \mathcal{S}\times\mathcal{A}\to \Delta(\mathcal{S})$ denotes the stochastic transition kernel such that $\mathcal{P}(\cdot \mid s,a)$ is a probability measure over next state given that the current state is \(s\) and joint action is \(a\); $\mathcal{Z}_i \subseteq \mathbb{R}^{d_{z_i}}$ denotes the (continuous) observation space of agent $i \in \mathcal{N}$ and $\mathcal{Z} = \times_{i\in \mathcal{N}} \mathcal{Z}_i$ denotes the joint observation space; $\Delta(\mathcal{Z}_i)$ denotes the set of probability measures on $\mathcal{Z}_i$; $\mathcal{O}_i: \mathcal{S}\times\mathcal{A}\to \Delta(\mathcal{Z}_i)$ denotes the stochastic observation kernel of agent $i$ such that $\mathcal{O}_i(\cdot\mid s, a)$ is probability measure over \(\mathcal{Z}_i\) given current state \(s\) and joint action \(a\); $r_i: \mathcal{S}\times\mathcal{A} \to \mathbb{R}$ denotes the stage reward function\footnote{Although we do not explicitly mention it, we assume standard measurability restrictions on various functions defined in this manuscript.} of agent $i$; $\gamma \in (0,1)$ denotes the discount factor (to be defined shortly); and $\rho \in \Delta(\mathcal{S})$ denotes the initial state distribution. 

The game proceeds in discrete time steps, indexed by \(t\in \mathbb{N}\). At any time \(t\), the state, action, and observation is denoted by \((s_t, (a_{i,t})_{i\in\mathcal{N}}, (o_{i,t})_{i\in\mathcal{N}}).\) The initial state is denoted \(s_0\sim\rho\). At any time \(t,\) agents take their action based on the history of information available to them. More concretely, let $\mathcal{H}_{i,t} \subset \mathcal{Z}^t \times \mathcal{A}^{t-1}$ denote the history of information available to agent \(i\). Given history \(h_{i,t}\in \mathcal{H}_{i,t}\), agent \(i\) takes action \(a_{i,t}\) according to a policy \(\pi_i:\mathcal{H}_{i,t}\rightarrow\Delta(\mathcal{A}_i)\) such that \(\pi_i(\cdot|h_{i,t})\) is a probability measure over \(\mathcal{A}_i\). The joint policy is defined as $\pi = (\pi_i)_{i\in \mathcal{N}}$. We use \(\Pi_i\) to denote the set of history-dependent policy space of player \(i\) and \(\Pi=\times_{i\in\mathcal{N}}\Pi_i\) to denote the joint policy space of all players.  

The goal of every agent \(i\) is to maximize its own utility, captured by expected long-run discounted utility  $J_i(\pi) = \mathbb{E}_{\mathcal{P}, \rho, \mathcal{O}, \pi}\left[\sum_{t=0}^\infty\gamma^tr_i(s_t, {a}_t)\right]$. Note that the utility function of agent \(i\) depends on not only its own policy but also the policy of other agents. This coupled interaction makes the game a significantly more challenging problem compared to single-agent optimization. Moreover, we do not impose any assumption on the function \(r_i\) (or for that matter \(J_i\)). The reward functions of agents could be aligned or conflicting. 

A common desideratum in POMGs is to compute Nash equilibrium, which is a joint policy where no agent has any incentive to unilaterally deviate. More concretely, 
\begin{definition}
    A joint policy $\pi^* = \{\pi_i^*\}_{i\in[n]}$ is an \(\epsilon\)-approximate Nash equilibrium, for some \(\epsilon\geq 0,\) if
\begin{equation}
\label{eq:ne_def}
J_i(\pi^*) \geq J_i(\pi'_i, \pi^*_{-i}) - \epsilon \quad \forall \pi'_i \in \Pi_i, \forall i \in \mathcal{N}.
\end{equation}
If \(\epsilon=0\), then \(\pi^\ast\) is simply a Nash equilibrium. 
\end{definition}

Computation of (approximate) Nash equilibria is a challenging problem even if agents have full observation \cite{daskalakis2009}. Indeed, many MARL algorithms are only guaranteed to compute equilibrium tractably in some restricted classes of games like zero-sum \cite{zhang2024survey}, cooperative \cite{yu2022surprising}, or potential games \cite{leonardos2021global}. But many real world multi-agent interactions in dynamic environments do not fall in any of these regimes. 

Recently, \cite{guo2025markov} introduced the framework of 
$\alpha$-potential games, which provides us with a framework to approximate Nash equilibrium in general-sum multi-agent games. Based on this, we introduce the notion of a Markov near-potential function below:
\begin{definition}\label{def:near_pot_def}
    A function \(\Phi:\Pi\rightarrow\mathbb{R}\) is a Markov near-potential function (MNPF) of game \(\mathcal{G}\) with an approximation parameter \(\alpha \geq 0\) if 
\begin{equation}
\label{eq:apg_def}
\max_{i\in \mathcal{N}}\max_{\pi_i,\pi_i'\in \Pi_i}\max_{\pi_{-i}\in\Pi_{-i}} \big|J_i(\pi'_i, \pi_{-i}) - J_i(\pi_i, \pi_{-i}) - (\Phi(\pi'_i, \pi_{-i}) - \Phi(\pi_i, \pi_{-i}))\big| \leq \alpha.
\end{equation}
\end{definition}

Definition \ref{def:near_pot_def}  demands that the difference between the change in value function of any player due to a unilateral change in its policy and the change in MNPF is at most \(\alpha\). Note that if there exists an MNPF associated with \(\alpha=0\), then the game is called a Markov potential game \cite{leonardos2021global}.

\begin{proposition}\label{prop:MNPF}
   Consider a game \(\tilde{\mathcal{G}}\), which shares the same structure as \(\mathcal{G}\), with the only difference being that all players in \(\tilde{\mathcal{G}}\) share the same utility function equal to an MNPF \(\Phi\) of \(\mathcal{G}\) with approximation parameter \(\alpha\). If \(\tilde{\pi}\in \Pi\) is a Nash equilibrium of \(\tilde{\mathcal{G}}\), then \(\tilde{\pi}\) will be an \(\alpha-\)approximate Nash equilibrium of the original game \(\mathcal{G}\). 
\end{proposition}
Proof of proposition \ref{prop:MNPF} is found in Appendix \ref{sec:ProofMNPF}. This shows that computing an \(\alpha\)-approximate equilibrium of game \(\mathcal{G}\) is possible by computing a Nash equilibrium of a cooperative game where utility of all agents is an MNPF of game \(\mathcal{G}\) with an approximate parameter \(\alpha.\)

Note that solving for Nash equilibria of cooperative Markov games is possible using many popular algorithms like HAPPO \cite{zhong2024heterogeneous}, MAPPO \cite{yu2022surprising}, etc. Also, simply maximizing MNPF yields a Nash equilibrium of a cooperative game \cite{guo2025markov}. 

To summarize, if we have access to a candidate MNPF with a low value of \(\alpha,\) we can closely approximate the Nash equilibrium of the original game. Using this insight, in the next section, we develop a MARL algorithm that \textit{(i)} learns a potential function candidate \(\Phi\) by minimizing a novel metric (to be introduced in the next section) that amounts to minimizing \(\alpha;\) and \textit{(ii)} computes a Nash equilibrium of a cooperative game with utility function \(\Phi.\)

\section{Near-Potential Policy Optimization Algorithm}
In this section, we introduce our algorithmic framework \textit{Near-Potential Policy Optimization}, which simultaneously learns a potential function candidate $\Phi$ and, using that, learns an approximate equilibrium policy for the original game. First, we introduce a novel optimization metric which provably ensures that minimizing this metric will give an MNPF candidate that closely approximates the underlying game. Next, using that novel optimization metric, we develop a novel MARL training pipeline that computes an approximate equilibrium of the original game. 

\subsection{Optimization Metric to Learn MNPF}
One natural approach to learning a Markov near-potential function (MNPF) is to directly minimize the deviation from the defining condition in \eqref{eq:apg_def}. Specifically, one could solve
\begin{align}
    \min_{\Phi}\max_{i\in \mathcal{N}}\max_{\pi_i,\pi_i'\in \Pi_i}&\max_{\pi_{-i}\in\Pi_{-i}}
    \big|\notag 
    J_i(\pi'_i, \pi_{-i}) - J_i(\pi_i, \pi_{-i}) - \big(\Phi(\pi'_i, \pi_{-i}) - \Phi(\pi_i, \pi_{-i})\big)
    \big|.
\end{align}
This objective, based on Definition \ref{def:near_pot_def}, seeks a potential function whose unilateral incentive differences match those of the underlying game across all policy profiles. However, solving this optimization problem is computationally challenging: the resulting objective is non-convex--non-concave, and such problems are known to be difficult to approximate even locally \cite{maheshwari2025exotic, lin2020gradient}. Moreover, enforcing the near-potential condition uniformly over all policy profiles may be unnecessarily restrictive when the ultimate goal is to compute a Nash equilibrium. 

Motivated by this observation, we develop a proxy objective that focuses on learning a potential function that accurately captures incentive differences around equilibrium-relevant policies, rather than across the entire policy space.
Intuitively, a key insight is that, in order to compute an approximate Nash equilibrium, we only need to approximate the changes in the utility function of players due to unilateral deviations around a reference policy. Importantly, this reference policy is a Nash equilibrium of a cooperative game with the MNPF as its utility function. More formally, we define a function 
{\small\begin{equation}
\label{eq:relaxed_F_i}
F_i(\Phi) = \Phi(\pi^{*,\Phi}) - \Phi(\pi_i^{*, J}, \pi^{*, \Phi}_{-i}) - (J_i(\pi^{*, \Phi}) - J_i(\pi_i^{*,J}, \pi_{-i}^{*, \Phi})),
\end{equation}}
\noindent where \(\pi^{*, \Phi}\) is such that for every \(i\in \mathcal{N},\) $\pi^{*, \Phi}_i \in \arg\max_{\pi_i \in \Pi_i}\Phi(\pi_i, \pi^{*, \Phi}_{-i})$. Furthermore, $\pi^{*, J}_i \in \arg\max_{\pi_i \in \Pi_i} J_i(\pi_i, \pi^{*, \Phi}_{-i})$ is the best response to player \(i\) when others are using policy  \(\pi^{*, \Phi}_{-i}\).\footnote{Note that ties are broken arbitrarily.} 
In words, $F_i(\Phi)$ is the difference between the change in \(\Phi\) due to unilateral shifts of player \(i\) to its best response at the Nash equilibrium \(\pi^{\ast,\Phi}\) and the change in  player \(i'\)s individual utility. Note, by \eqref{eq:relaxed_F_i}, \(F_i\) is non-negative for all players \(i\in \mathcal{N}\). Under mild conditions, it can be shown that \(F_i\) is upper-hemicontinuous.\footnote{Because the equilibrium $\pi^{*,\Phi}$ need not be unique, the mapping $F_i$ should formally be viewed as a \emph{correspondence} (i.e. set-valued mappings). Upper hemicontinuity \cite{aliprantis2006infinite} provides a standard regularity condition ensuring stability of attainable values under perturbations of $\Phi$. In our implementation, the reinforcement learning procedure used to compute $\pi^{*,\Phi}$ acts as an implicit selection rule from this correspondence, producing a single realized value of $F_i(\Phi)$ during training.} 

Next, we show the usefulness of \(F_{i}\) as a metric to compute approximate Nash equilibrium of original game. 
\begin{theorem}
\label{thm:ApproxNashThroughPotentialNew}
Consider a function $\Phi:\Pi\rightarrow\mathbb{R}$ such that $\max_iF_i(\Phi) \leq \alpha$. Then, $\pi^{*, \Phi}$ is an $\alpha$-Nash equilibrium.
\end{theorem}
A proof of Theorem \ref{thm:ApproxNashThroughPotentialNew} is provided in Appendix \ref{sec:proofAltFi}. Theorem \ref{thm:ApproxNashThroughPotentialNew} ensures that minimizing the maximum value of \(F_i\) across players is enough to approximate a Nash equilibrium. This property will be the backbone of our algorithmic methodology described in next section.

Next, we highlight another structural property of the objective \(F_i(\cdot)\), and its connection to potential games \cite{leonardos2021global}.
\begin{proposition}\label{prop:FiEqualsZero}
If a game \(\mathcal{G}\) is a Markov potential game with potential function \(\Phi\), then \(F_i(\Phi)=0\) for all \(i\in\mathcal{N}\). 
However, there exist games that are not potential games, but there exists \(\Phi\) such that \(F_i(\Phi)=0\) for all \(i\in\mathcal{N}\).
\end{proposition}

A proof of Proposition \ref{prop:FiEqualsZero} is provided in Appendix \ref{sec:ProofFiEqualsZero}. Proposition \ref{prop:FiEqualsZero} establishes that having \(F_i(\Phi)=0\) for all \(i\in \mathcal{N}\) is only a necessary condition for potential games and not a sufficient condition. 

\subsection{MARL Algorithm to Approximate Nash Equilibria in General-Sum Games}
Using the insight from Theorem~\ref{thm:ApproxNashThroughPotentialNew}, we introduce a MARL pipeline that aims to solve
\begin{align}\label{eq:OriginalEq}
\min_{\Phi:\Pi\rightarrow\mathbb{R}}F(\Phi) = \max_{i\in \mathcal{N}} F_i(\Phi),
\end{align}
while simultaneously generating the policy \(\pi^{*,\Phi}\). As written, this is an infinite-dimensional non-smooth optimization problem. The non-smoothness arises from the inner maximization \(\max_{i\in \mathcal{N}}F_i(\Phi)\), while the outer minimization is taken over the space of real-valued functions on \(\Pi\). To make this problem tractable, we introduce two modifications.

First, instead of optimizing over all real-valued functions on \(\Pi\), we restrict the candidate potential function to a parameterized family. Specifically, let \(w\in \mathcal{W}\subset \mathbb{R}^p\) be a parameter vector. For any such \(w\), we define a parameterized function
\(\Phi_w:\Pi\rightarrow\mathbb{R}\).\footnote{In practical applications, a suitable choice of \(\Phi_w = Z(x) +\psi_w(x)\) where \(Z(x)\) is certain system objective and \(\psi_w(x)\) is certain low-dimensional adaptation to it.} 

Second, we address the non-smoothness introduced by the maximization over agents by replacing \(\max_{i\in\mathcal{N}}F_i(\Phi)\) with the following smooth approximation:
\[
\tilde{F}_\beta(\Phi) :=
\frac{1}{\beta}\log\!\left(
\sum_{i\in\mathcal{N}}
\exp\!\left(\beta F_i(\Phi)\right)
\right).
\]
It is well known that \(
\left|
\max_{i\in\mathcal{N}}F_i(\Phi)
-
\tilde{F}_\beta(\Phi)
\right|
\le
\frac{\log(|\mathcal{N}|)}{\beta},
\)
which implies that the smoothed objective approximates the original objective arbitrarily well for sufficiently large \(\beta\).
Therefore, in the remainder of this section we focus on solving the following optimization problem:
\begin{align}\label{eq:mainMARL}
\min_{w\in\mathcal{W}} \tilde{F}_\beta(\Phi_w).
\end{align}
To solve \eqref{eq:mainMARL}, we propose Algorithm~\ref{alg:nppg}, which consists of three main modules.

\textsf{(M1)} \textbf{Gradient-descent module:}  
We employ gradient descent to solve \eqref{eq:mainMARL}. However, computing the gradient of \(\tilde{F}_{\beta}(\cdot)\) requires evaluating the gradient of \(F_i(\cdot)\) for every player \(i\in \mathcal{N}\), which is challenging for two reasons. First, from \eqref{eq:relaxed_F_i}, we observe that \(F_i(\cdot)\) depends on \(\Phi\) through \(\pi^{*,\Phi}\), which is the Nash equilibrium of a cooperative game where \(\Phi\) acts as the common utility function. Second, \(F_i(\cdot)\) depends on \(\Phi\) through \(\pi_i^{*,J}\), which is the best response of player \(i\) when other players use the strategy \(\pi_{-i}^{*,\Phi}\). Since \(\pi_{-i}^{*,\Phi}\) itself depends on \(\Phi\), the resulting dependence creates a bilevel optimization structure. This makes backpropagation-based gradient computation difficult.

To circumvent this issue, we employ a zeroth-order gradient approximation. Specifically, we estimate the gradient of \(\tilde{F}_{\beta}(\Phi_w)\) at \(w\) by sampling a random direction \(u\) uniformly from the unit sphere in \(\mathbb{R}^p\), and computing the following two-point estimator:
\begin{equation}
\label{eq:zeroth-order-approx}
\hat\nabla_w \tilde{F}_{\beta}(w;\delta)
=
\frac{p}{2\delta}
\left(
\tilde{F}_\beta(\Phi_{\hat w})
-
\tilde{F}_\beta(\Phi_{\check w})
\right)u,
\end{equation}
where \(\hat{w}= w+\delta u\) and \(\check{w} = w-\delta u\). 
The bias of this estimator decreases as \(\delta \rightarrow 0\), at the cost of increased variance \cite{flaxman2004online}. The choice of \(\delta\) must therefore be balanced with the step size of the gradient-descent update.

\textsf{(M2)} \textbf{Estimating \(\tilde{F}_\beta(\Phi_w)\):}  
To compute \eqref{eq:zeroth-order-approx}, we must evaluate \(\tilde{F}_\beta(\Phi_w)\) at two parameter values \(w\). Recall that computing \(\tilde{F}_\beta(\Phi_w)\) requires evaluating \(F_i(\Phi_w)\) for every player \(i\in \mathcal{N}\), as defined in \eqref{eq:relaxed_F_i}. This requires computing three objects: 
(i) \(\pi^{*,\Phi_w}\), 
(ii) \(\pi_i^{*,J}\), and 
(iii) evaluating value functions associated with the resulting policies.

\textit{(i) Computing \(\pi^{*,\Phi_w}\):}  
Recall that \(\pi^{*,\Phi_w}\) is the Nash equilibrium of a cooperative game in which every player maximizes the shared objective \(\Phi_w\). 
This step corresponds to the \textsc{CoopGameSolver} module in Algorithm~\ref{alg:nppg}.

\textit{(ii) Computing \(\pi_i^{*,J}\) for each player \(i\in \mathcal{N}\):}  
The policy \(\pi_i^{*,J}\) denotes the best response of player \(i\) when all other players follow \(\pi_{-i}^{*,\Phi_w}\). Computing this best response reduces to a standard reinforcement learning problem. 
This step corresponds to the \textsc{RLSolver} module in Algorithm~\ref{alg:nppg}.

\textit{(iii) Evaluating value functions:}  
Given \(\pi^{*,\Phi_w}\) and \(\{\pi_i^{*,J}\}_{i\in\mathcal{N}}\), we estimate the quantities \\
\noindent\(
\Phi_w(\pi^{*,\Phi_w}), ~
\Phi_w(\pi_i^{*,J},\pi_{-i}^{*,\Phi_w}), ~
J_i(\pi^{*,\Phi_w}), ~
J_i(\pi_i^{*,J},\pi_{-i}^{*,\Phi_w})\)
using Monte-Carlo rollouts. These estimates allow us to compute \(F_i(\Phi_w)\) and consequently \(\tilde{F}_\beta(\Phi_w)\).

\paragraph{Description of Algorithm~\ref{alg:nppg}}
Algorithm~\ref{alg:nppg} iteratively refines estimates of the potential function, the Nash equilibrium of the cooperative game with the potential as the common utility (through \(\pi^{\Phi}\)), and the best responses of individual players (through \(\{\pi_i^{J}\}_{i\in\mathcal{N}}\)).

At each iteration, the algorithm samples a random direction \(u\) uniformly from the unit sphere in \(\mathbb{R}^p\), and uses it to construct perturbed parameter vectors \(\hat{w}\) and \(\check{w}\) (Line~4 of Algorithm~\ref{alg:nppg}). These perturbed parameters are used to compute the two-point zeroth-order gradient estimate.

We then estimate the Nash equilibrium of the cooperative games induced by the potential functions corresponding to the perturbed parameters. In particular, we compute approximate equilibria of the cooperative games with parameters \({\hat{w}}\) and \({\check{w}}\), respectively (Lines~5--6 of Algorithm~\ref{alg:nppg}). This step is implemented using the \textsc{CoopGameSolver} module, which takes the current policy estimate \(\pi^{\Phi}\) (used as a warm start) 
and performs \(K_1\) number of updates using cooperative MARL algorithms \cite{zhong2024heterogeneous,yu2022surprising}. The number of solver iterations is treated as a hyperparameter and must be tuned in practice.

We then compute approximate best responses of each player to the strategies returned by \textsc{CoopGameSolver}. This reduces to a single-agent reinforcement learning problem and is handled by the \textsc{RLSolver} module (Lines~8--9 of Algorithm~\ref{alg:nppg}). The \textsc{RLSolver} module takes as input the current estimate of the best-response policy \(\pi_i^{J}\) (used as a warm start), the policies of other agents returned by the cooperative game solver, the one-stage reward function of player \(i\), and performs \(K_2\) number of policy updates. 

Using the resulting policy estimates, we evaluate \(\tilde{F}_\beta(\Phi_{\hat{w}})\) and \(\tilde{F}_{\beta}(\Phi_{\check{w}})\) using Monte-Carlo estimation (Lines~11--14 of Algorithm~\ref{alg:nppg}). Finally, we update the parameter vector \(w\) using the zeroth-order gradient step (Line~15 of Algorithm~\ref{alg:nppg}). This process is repeated for multiple iterations until convergence. 

\begin{algorithm}
    \caption{Near-Potential Policy Optimization}
    \label{alg:nppg}
    \begin{algorithmic}[1]
        \State \textbf{Input:} The zeroth-order step size parameter \(\delta \geq 0\), gradient descent learning rate \(\eta \geq 0\), and smoothness parameter \(\beta \geq 0\).
        \State Initialize \(\pi^{\Phi}\), \(\pi^{J}_i\) for all agents \(i \in [n]\), and \(w \in \mathcal{W}\)
        \For{each iteration}
            \State Sample \(u \sim \mathcal{S}(\mathbb{R}^{\dim(w)})\)
            \State Set \(\hat{w} = w + \delta u\) and \(\check{w} = w - \delta u\)
            \State \(\check\pi^\Phi \leftarrow \textsc{CoopGameSolver}(\pi^{\Phi}, {\check w}, K_1)\)
            \State \({\pi}^\Phi \leftarrow \textsc{CoopGameSolver}(\pi^{\Phi}, {\hat w}, K_1)\)
            \For{each agent \(i \in \mathcal{N}\)}
                \State \(\check \pi^J_i \leftarrow \textsc{RLSolver}(\pi^{J}_i, \check\pi^{\Phi}_{-i}, r_i, K_2)\)
                \State \(\pi^J_i \leftarrow \textsc{RLSolver}(\pi^{J}_i, \pi^{\Phi}_{-i}, r_i, K_2)\)
            \EndFor
            \For{each agent \(i \in \mathcal{N}\)}
                \State Compute \({F}_i(\Phi_{\hat w})\) and \({F}_i(\Phi_{\check w})\) as per \eqref{eq:relaxed_F_i}
            \EndFor
            \State Set \(\tilde F_\beta(\Phi_{\hat w})= \frac{1}{\beta}\log\!\left(\sum_{i} \exp(\beta {F}_{i}(\Phi_{\hat w}))\right)\)
            \State Set \(\tilde F_\beta(\Phi_{\check w})= \frac{1}{\beta}\log\!\left(\sum_{i} \exp(\beta {F}_{i}(\Phi_{\check w}))\right)\)
            \State \(w \leftarrow w - \eta\frac{\dim(w)}{2\delta}\left(\tilde F_\beta(\Phi_{\hat w}) - \tilde F_\beta(\Phi_{\check w})\right)u\)
        \EndFor
        \State \Return \(\pi^\Phi\)
    \end{algorithmic}
\end{algorithm}

\section{Experimental Evaluation}
In this section, we evaluate the proposed algorithm on various examples. 
We begin with a class of two-player, two-action matrix games to illustrate the mechanism of Algorithm~\ref{alg:nppg} and compare it with standard policy-gradient baselines such as IPPO and MAPPO. 
We then study a general-sum partially observable dynamic game based on the \emph{Simple World Comm} benchmark from the Multi-Particle Environment suite. 

Unless stated otherwise, throughout this section we instantiate $\textsc{CoopGameSolver}$ and $\textsc{RLSolver}$ in Algorithm~\ref{alg:nppg} using Heterogeneous-Agent Proximal Policy Optimization (HAPPO)~\cite{zhong2024heterogeneous} and PPO~\cite{schulman2017proximal}, respectively.

\subsection{Illustrating the mechanism of Algorithm~\ref{alg:nppg} through toy example}
We first use a matrix game to explain the role of Algorithm~\ref{alg:nppg}. 
The objective here is not to demonstrate large-scale performance, but to show concretely how the algorithm searches over a class of candidate potential functions.

Consider the following class of two-player, two-action matrix games parameterized by $\alpha \in [0,1]$:
\begin{equation}\label{eq:matrixgame_alpha24}
\begin{aligned}
\begin{array}{c|cc}
 & B_1 & B_2 \\ \hline
A_1 & (1,1-2\alpha) & (1-2\alpha,\frac{1}{2}(\alpha+1)) \\
A_2 & ((1-3\alpha)/2,(7-3\alpha)/4) & (\alpha, 2-3\alpha)
\end{array}
\end{aligned}
\end{equation}
At $\alpha=0$, the game is a general-sum game, whereas at $\alpha=1$ it becomes zero-sum.

To illustrate the algorithm, we first focus on the case $\alpha=0$, for which the game reduces to
\begin{align}\label{eq:matrxigame}
\mathcal{G} &= \begin{array}{c|cc}
 & B_1 & B_2 \\ \hline
A_1 & (1,\, 1) & (1,\, 1/2) \\
A_2 & (1/2,\, 7/4) & (0,\, 2)
\end{array}.
\end{align}

For this game, we parameterize the candidate potential function as a convex combination of the two players' utilities: \(
\Phi_w = w J_1 + (1-w)J_2\) such that \(w\in [0,1].\)
This yields the payoff matrix
\begin{align*}
\Phi_w=
\begin{array}{c|cc}
 & B_1 & B_2 \\ \hline
A_1 & 1 & (1+w)/2 \\
A_2 & (7-5w)/4 & 2(1-w)
\end{array}.
\end{align*}

Thus, in this example, the role of Algorithm~\ref{alg:nppg} is to select a value of $w$ such that a Nash equilibrium of the cooperative game induced by $\Phi_w$ is also an approximate Nash equilibrium of the original game. 
The objective in \eqref{eq:OriginalEq} can be explicitly evaluated, as stated next.

\begin{proposition}\label{prop:toy_example}
The game in \eqref{eq:matrxigame} satisfies
\begin{align*}
F(\Phi_w)=&\max\{F_1(\Phi_w), F_2(\Phi_w)\}\\ 
        &= \begin{cases}
        5\frac{1-w}{2}, & \text{if } w < 1/3,\\
        \{\frac{5}{6}(2-\alpha) : \alpha \in [0,1]\},& \text{if } w = 1/3,\\
        5\frac{1-w}{4},& \text{if } w \in (1/3,0.6),\\
        \{\frac{1-\beta}{2}:\beta\in[0,1]\},& \text{if } w = 0.6, 
        \\ 
        0, & \text{if } w \in (0.6,1]. 
    \end{cases}
\end{align*}
Furthermore, $F(\Phi_w)$ is minimized over all $w \in [0.6,1]$, and its minimum value is $0$.
\end{proposition}

The proof of Proposition \ref{prop:toy_example} is in Appendix \ref{sec:toy_example_proof}. Proposition~\ref{prop:toy_example} shows that although the game in \eqref{eq:matrxigame} is not a potential game (see Appendix \ref{sec:NotPotentialProof}), there exists a potential function within the class $\{\Phi_w\}_{w\in[0,1]}$ such that a Nash equilibrium of the induced cooperative game coincides with a Nash equilibrium of the original game.

\begin{figure}[t]
    \centering

    \begin{minipage}{0.48\linewidth}
        \centering
        \includegraphics[width=\linewidth]{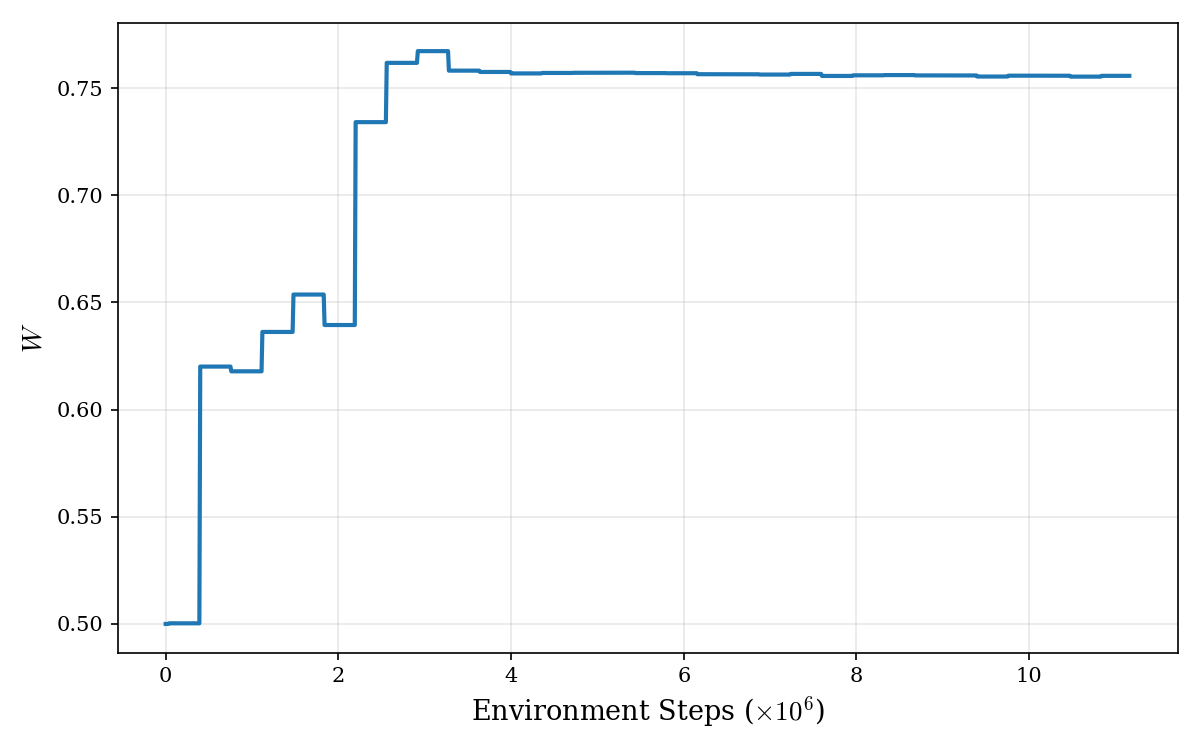}
        
        \small (a) Evolution of \(w\).
    \end{minipage}
    \hfill
    \begin{minipage}{0.48\linewidth}
        \centering
        \includegraphics[width=\linewidth]{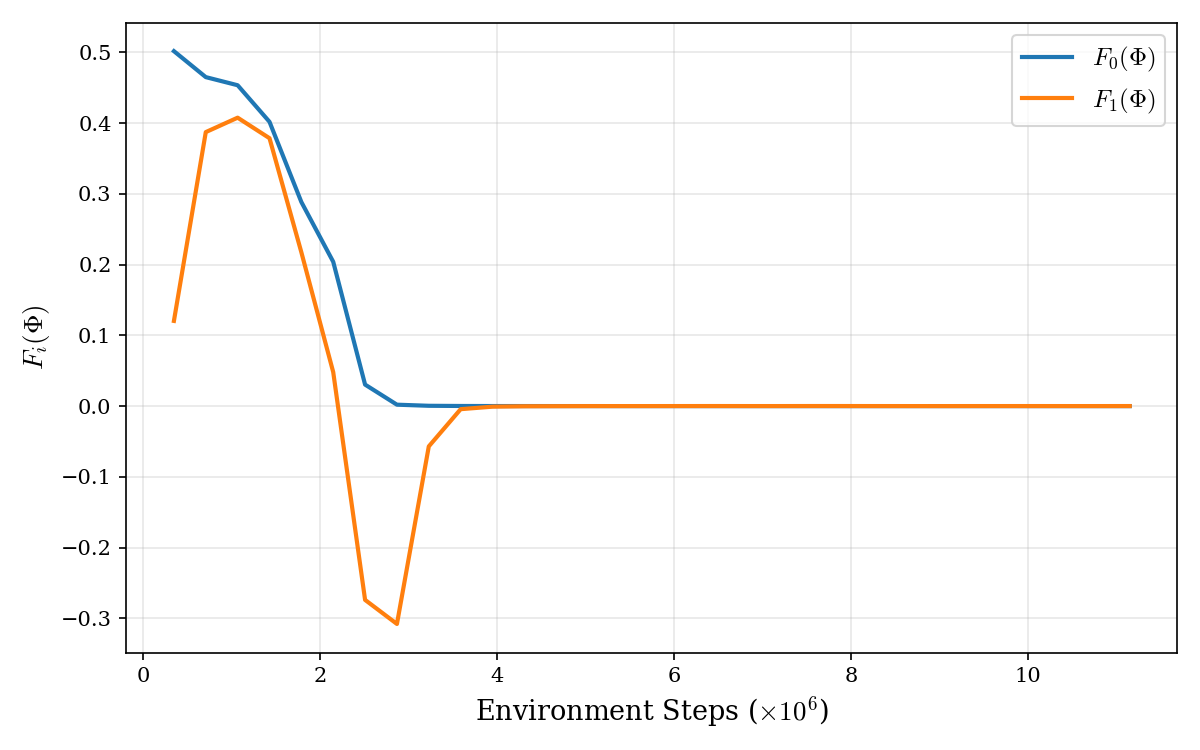}
        
        \small (b) Evolution of \(F_i(\Phi_w)\).
    \end{minipage}

    \vspace{0.3cm}

    \begin{minipage}{0.48\linewidth}
        \centering
        \includegraphics[width=\linewidth]{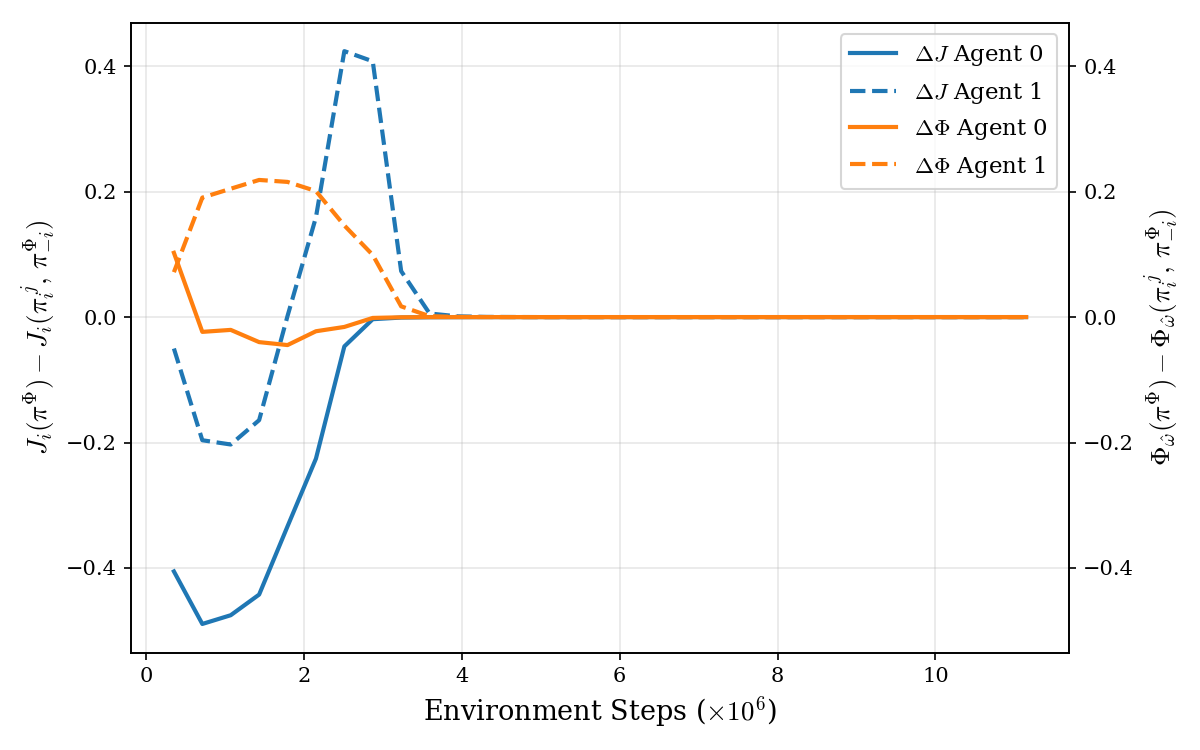}
        
        \small (c) Evolution of value and potential changes.
    \end{minipage}
    \hfill
    \begin{minipage}{0.48\linewidth}
        \centering
        \includegraphics[width=\linewidth]{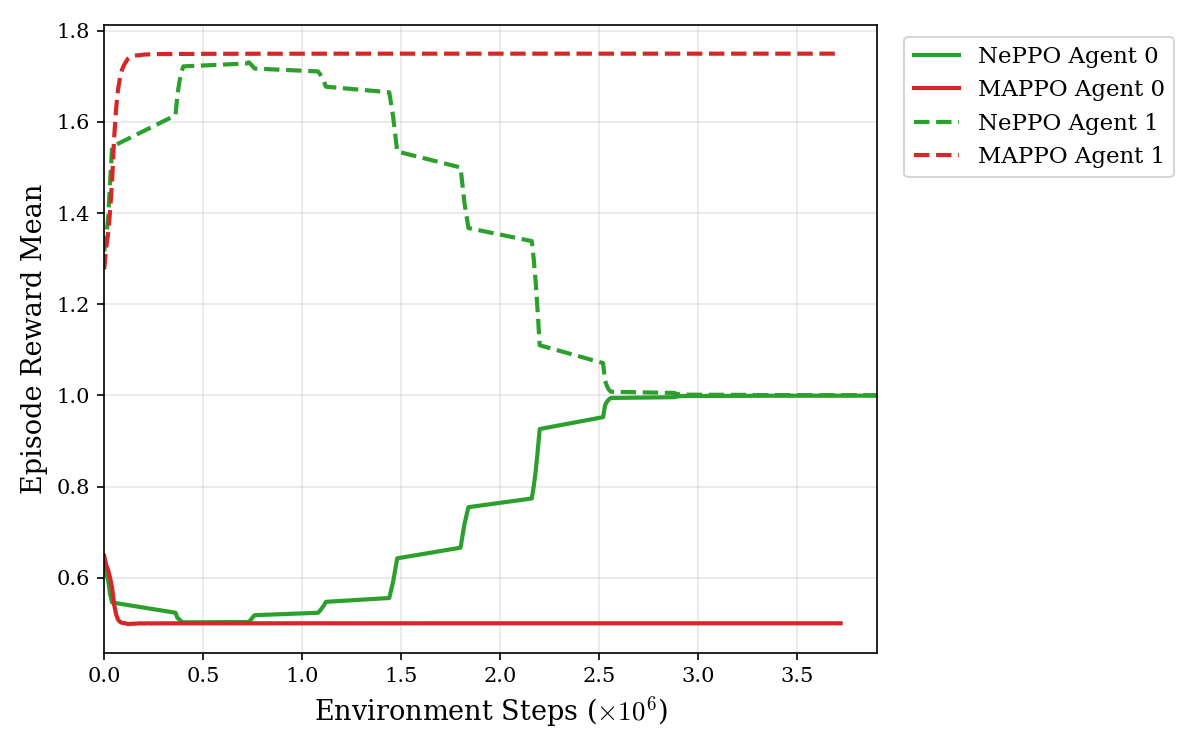}
        
        \small (d) MAPPO versus Algorithm \ref{alg:nppg}.
    \end{minipage}

    \caption{Toy-example results for the game in \eqref{eq:matrxigame}. Panel (a) shows the evolution of \(w\) in Algorithm \ref{alg:nppg}. Panel (b) shows the evolution of \(F_i(\Phi_w)\). Panel (c) shows the change in value function and potential function due to unilateral best responses. Panel (d) compares MAPPO with Algorithm \ref{alg:nppg}.}
    \label{fig:toy_example_results}
\end{figure}

Next, we show that our algorithm is indeed able to recover the Nash equilibrium.
In Figure \ref{fig:toy_example_results}(a), we see the evolution \(w\) in one run of Algorithm \ref{alg:nppg} for the game in \eqref{eq:matrxigame}. We see that \(w\) approaches to around \(0.75\), which, as per Proposition \ref{prop:toy_example}, is a minimizer of \(\max\{F_1(\Phi_w), F_2(\Phi_w)\}.\) This is also evidenced by Figure \ref{fig:toy_example_results}(b). Next, from Figure \ref{fig:toy_example_results}(c), we note that the change in players' value function due to unilateral best response and the change in potential function both approach \(0\). In the transient phase, we see that the change in the value function is positive (which violates the definition of best response), but this is because, in the transient, the best response module (i.e. \textsc{RLSolver}) has not converged to an optimal solution. Similarly, the change in potential is negative in some instances. This is because the \textsc{CoopGameSolver} has not converged by then. We see however that these solvers are eventually able to produce the correct optimizers. Finally, in Figure \ref{fig:toy_example_results}(d), we compare the performance of our algorithm with SOTA algorithms such as MAPPO. Since MAPPO optimizes the function \(0.5*J_1+0.5*J_2\), it converges to the action \((A_2, B_1)\), which is not a Nash equilibrium in \eqref{eq:matrxigame}. Meanwhile, our algorithm recovers the right Nash equilibrium, which has utility of \((1,1)\) for the players. This behavior is similarly seen in more complex environments, as discussed next. 

\vspace{0.4cm}
\subsection{Comparison with Baseline Algorithms}
We now compare Algorithm~\ref{alg:nppg} (NePPO) with IPPO and MAPPO on the class of games defined in \eqref{eq:matrixgame_alpha24}. 
We evaluate performance for $\alpha \in \{0, 0.2, 0.4, 0.6, 0.8, 1\}$.
For this experiment, we parameterize the potential function as \(
\Phi_w(x, y) = -(x - p)^2 - (y - q)^2,\)
where $x$ and $y$ denote the probabilities of player~1 choosing $A_2$ and player~2 choosing $B_2$, respectively, and $w=(p,q)\in[0,1]^2$.

Table~\ref{tab:alpha_max_regret} reports the maximum regret achieved by each algorithm. 
We observe that NePPO consistently achieves the lowest regret across all values of $\alpha$, and in several cases recovers exact Nash equilibria.

\begin{table}[h]
\centering
\caption{Max-Regret for Varying $\alpha$ by Algorithm}
\label{tab:alpha_max_regret}
\begin{tabular}{c|cccc}
\hline
$\alpha$ & NePPO & IPPO & MAPPO \\
\hline
0.0 & \textbf{0.000} & \textbf{0.000} & 0.500 \\
0.2 & \textbf{0.000} & \textbf{0.000} & 0.800 \\
0.4 & \textbf{0.036} & DNC & 1.100 \\
0.6 & \textbf{0.052} & DNC & 1.400 \\
0.8 & \textbf{0.034} & DNC & 1.700 \\
1.0 & \textbf{0.000} & 0.011 & N/A \\
\hline
\end{tabular}
\end{table}

\paragraph{Performance in zero-sum regimes}
A notable observation is that NePPO successfully recovers a Nash equilibrium even in the zero-sum case ($\alpha=1$). 
This is in contrast to the classical limitation that potential games cannot represent zero-sum interactions~\cite{monderer1996potential, xu2023fictitious}. 
The key reason is that NePPO does not rely on a global potential structure. Instead, it constructs a local potential approximation that matches unilateral deviations in a neighborhood of equilibrium, which is sufficient for recovering Nash strategies.

\begin{figure}[t] 
    \centering
    \includegraphics[width=\linewidth]{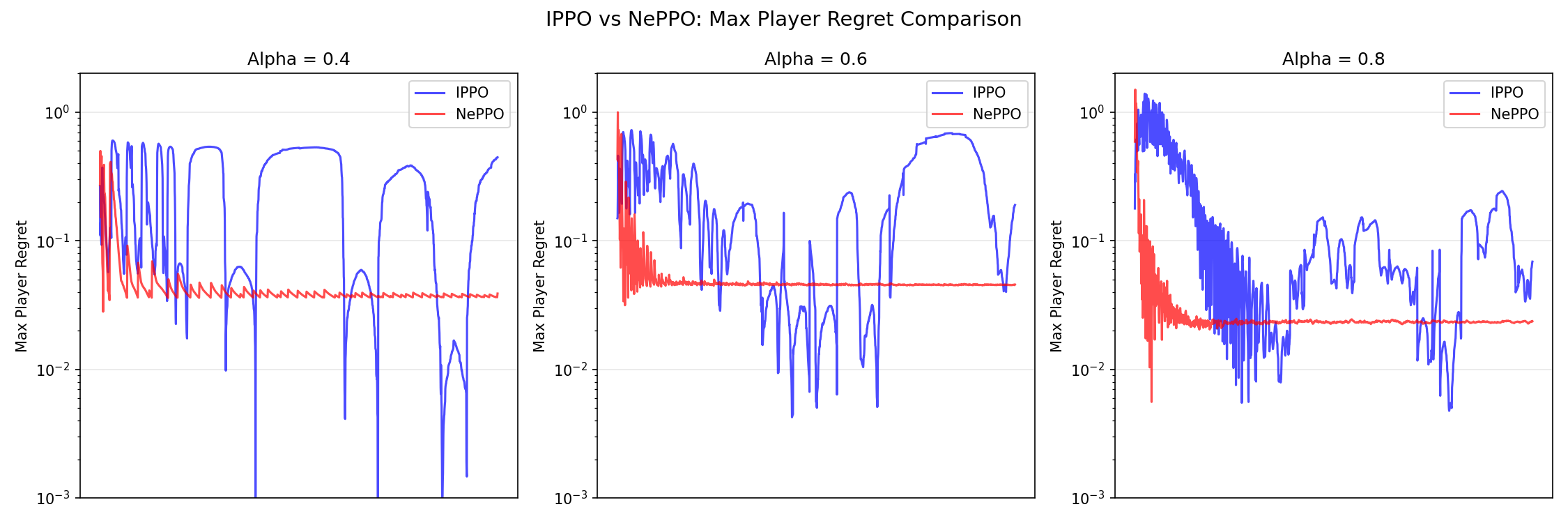}
\caption{Maximum player regret over training at $\alpha \in \{0.4, 0.6, 0.8\}$. NePPO runs for 500 iterations, whereas IPPO runs for 5000 iterations.}
\label{fig:ippo_conv}
\end{figure}

\paragraph{Failure modes of IPPO and MAPPO}
We observe that IPPO fails to converge (DNC) for $\alpha \in \{0.4, 0.6, 0.8\}$. 
Figure~\ref{fig:ippo_conv} shows that IPPO exhibits cycling behavior, whereas NePPO converges to a low-regret solution.

MAPPO, on the other hand, does not target Nash equilibria and instead optimizes a shared cumulative objective. 
As a result, it fails to recover equilibrium strategies and incurs high regret across all settings. 
In particular, for $\alpha=1$, all joint strategies yield zero cumulative reward, causing MAPPO to remain near its initialization.

\subsection{Performance in Partially Observable Dynamic Game}

We evaluate Algorithm~\ref{alg:nppg} on the \emph{Simple World Comm} environment from the Multi-Particle Environment suite~\cite{lowe2017multi}. 
This environment is a partially observable, general-sum Markov game with both cooperative and competitive interactions.

\paragraph{Environment description}
The environment consists of six agents. There are two types of agents: \emph{heroes} and \emph{adversaries}. 
Hero agents aim to collect food while avoiding being tagged, whereas adversaries aim to tag heroes. 
This induces a general-sum interaction: heroes balance survival and reward collection, while adversaries focus solely on pursuit.

The environment also includes partial observability and communication. 
A designated lead adversary has full visibility of the heroes and can communicate information to other adversaries, making coordination and communication learning essential for performance.

\paragraph{Potential function parameterization}
We parameterize the potential function as an infinite-horizon discounted sum of a learned stage reward $\phi_w: \mathcal{S} \times \mathcal{A} \to \mathbb{R}$. 
The stage reward is defined as a softmax-weighted combination of individual agent rewards:
\begin{equation}
\label{eq:phi_param}
    \phi_w(s_t, a_t) = \sum_{i \in \mathcal{N}} \sigma_i(W s_t + b)\, r_i(s_t, a_t),
\end{equation}
where $\sigma(\cdot)$ denotes the softmax function, and $w = (W,b)$ are the parameters of a single-layer neural network with $W \in \mathbb{R}^{N \times d_s}$ and $b \in \mathbb{R}^N$.

This parameterization allows the potential function to adaptively weight agents’ rewards as a function of the state, enabling the representation of both cooperative and competitive interactions.

\paragraph{Evaluation protocol}
We compare NePPO against IPPO, MAPPO, and MADDPG with the objective of minimizing Nash regret. 
For each trained policy, we estimate regret by computing a best response using PPO while keeping other agents’ policies fixed.

Table~\ref{tab:world_comm_max_reg} reports the maximum regret achieved by each algorithm. 
NePPO significantly outperforms all baselines.

\begin{table}[h]
\centering
\caption{Max Regret for Environment by Algorithm}
\label{tab:world_comm_max_reg}
\begin{tabular}{c|ccc}
\hline
 & NePPO & IPPO & MAPPO \\
\hline
Simple World Comm & \textbf{11.14} & 23.90 & 51.78 \\
\hline
\end{tabular}
\end{table}

MAPPO optimizes a shared cumulative reward and therefore tends to favor one group of agents at the expense of others, resulting in high regret in general-sum settings. 
IPPO, which optimizes each agent independently, performs better in competitive aspects but struggles to learn coordinated behaviors required in this environment.

In contrast, NePPO explicitly targets regret minimization by aligning local unilateral deviations through the learned potential function. 
This enables it to balance competitive and cooperative objectives and achieve substantially lower regret.

We were not able to obtain stable convergence for MADDPG in this environment.

\begin{figure}[t] 
    \centering
    \includegraphics[width=\linewidth]{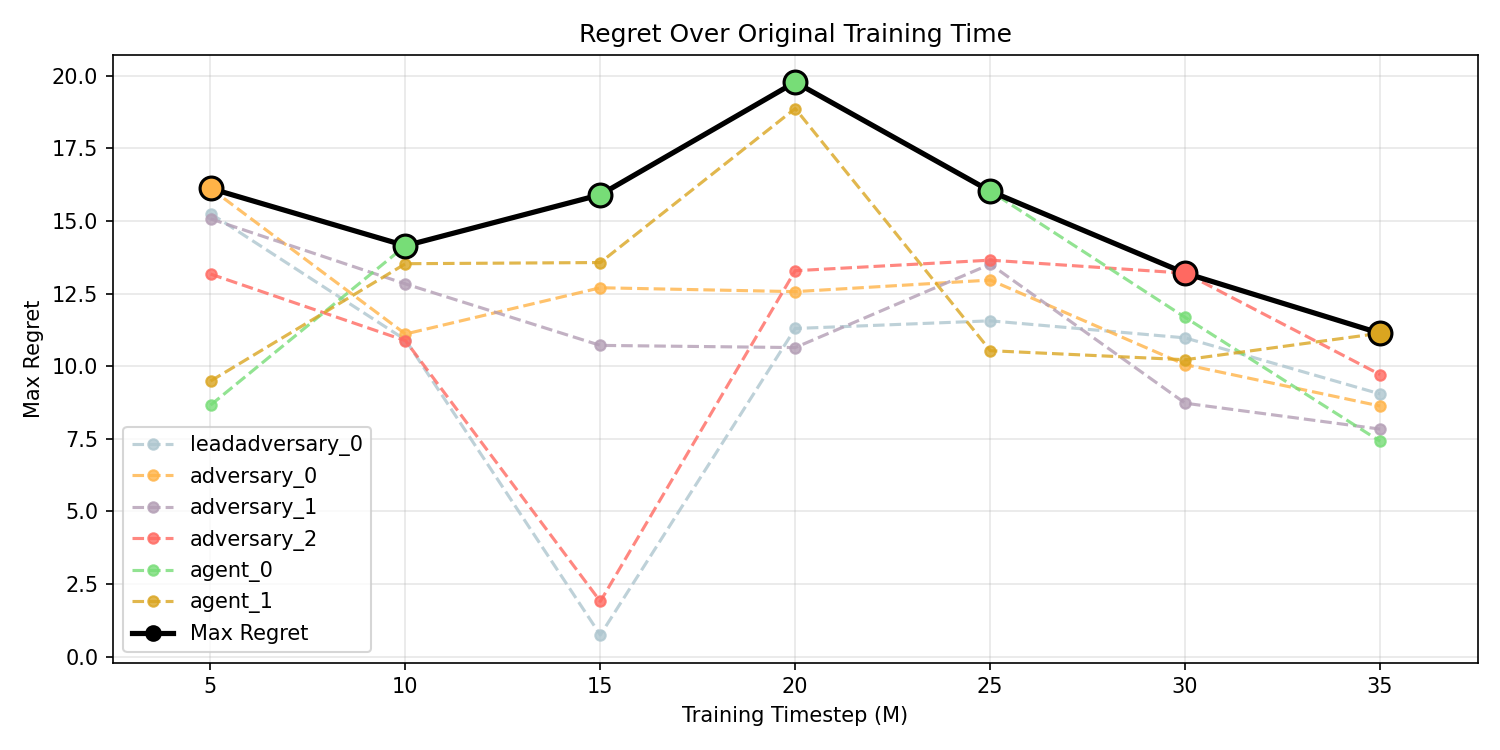}
\caption{Evolution of maximum regret across players over training steps.}
\label{fig:regret_mpe}
\end{figure}

\paragraph{Regret evolution over training}
Figure \ref{fig:regret_mpe} shows NePPO reduces the max regret over training in Simple World Comm, where it takes only one iteration of \textsc{CoopGameSolver} and \textsc{RLSolver} in every step of Algorithm \ref{alg:nppg}. Since it takes some steps for \textsc{CoopGameSolver} and \textsc{RLSolver} to have good performance, the regret initially increases before starting to decrease.

\section{Conclusions}
In this paper, we proposed a new MARL algorithm for approximating Nash equilibria in general-sum games. 
The key idea is to learn a shared potential function that captures changes in players' utilities under unilateral best-response deviations. 
This enables direct optimization of equilibrium behavior, rather than surrogate objectives such as cumulative reward. We showed that the proposed approach achieves low-regret approximate Nash equilibria across a range of settings, including zero-sum games where no global potential function exists. 

Several directions remain for future work. 
First, establishing convergence guarantees for the training dynamics of Algorithm~\ref{alg:nppg} would provide a theoretical foundation for the choice of cooperative game solver and RL solver. 
Second, developing first-order gradient estimators could reduce reliance on zeroth-order methods and improve sample efficiency. 
Finally, designing principled parameterizations of the potential function $\Phi$ is an important direction for improving both performance and scalability, serving as a design handle for equilibrium selection and training performance.

\bibliographystyle{plainnat}
\bibliography{main}

\appendix

\section{Proof of Proposition \ref{prop:MNPF}} %2.3
\label{sec:ProofMNPF}
Using the definition of $\Phi$, we know that for every $i \in \mathcal{N}$,
\begin{align*}
    |J_i(\pi'_i, \pi_{-i}) - J_i(\pi''_i, \pi_{-i}) - (\Phi(\pi'_i, \pi_{-i}) - \Phi(\pi''_i, \pi_{-i}))| \leq \alpha \quad \forall\pi'_i, \pi''_i \in \Pi_i, \forall\pi_{-i} \in \Pi_{-i}.
\end{align*}
Furthermore, as \(\tilde{\pi}\) is a Nash equilibrium of \(\tilde{G}\), for all agents $i \in \mathcal{N}$,
\begin{align}
    \label{eq:NashTildeG}
    \Phi(\tilde \pi) \geq \Phi(\pi'_i, \tilde \pi_{-i}) \quad  \forall\pi'_i \in \Pi_i. 
\end{align}
Using previous two inequalities, we note that for any \(\pi_{i}'\in \Pi_i,\)
\begin{align*}
J_i(\tilde \pi) - J_i(\pi'_i, \tilde \pi_{-i}) \geq \Phi(\tilde \pi) - \Phi(\pi'_i, \tilde \pi_{-i}) -\alpha \geq -\alpha. 
\end{align*} 
Thus, using the definition of approximate Nash equilibrium, we conclude that \(\tilde \pi\) is an \(\alpha-\)approximate Nash equilibrium of \(\mathcal{G}\).

\section{Proof of Theorem \ref{thm:ApproxNashThroughPotentialNew}}
\label{sec:proofAltFi}
To show that \({\pi}^{*,\Phi}\) is an \(\alpha-\)Nash equilibrium, we must establish that 
\begin{align*}
    J_i({\pi}^{*,\Phi}) - J_i(\pi_i^{*, J}, {\pi}^{*,\Phi}_{-i}) \geq -\alpha,
\end{align*}
where \(\pi_i^{*, J}\in \arg\max_{\pi_i\in\Pi_i}J_i(\pi_i, {\pi}^{*, \Phi}_{-i})\). Indeed, from \eqref{eq:relaxed_F_i}, we know that 
\begin{align*}
     J_i({\pi}^{*,\Phi}) - J_i(\pi_i^{*,J}, \bar{\pi}^{\Phi}_{-i}) &=  \Phi({\pi}^{*, \Phi}) - \Phi(\pi_i^{*,J}, {\pi}^{*,\Phi}_{-i}) - F_i(\Phi)
    \\
    &\geq 0-\alpha,
\end{align*}
where the last inequality is because of the fact that \(\Phi({\pi}^{*,\Phi}) - \Phi(\pi_i^{*, J}, {\pi}^{*,\Phi}_{-i})\geq 0\) as \({\pi}^{*,\Phi}\) is a Nash equilibrium of cooperative game with \(\Phi\) as players' utility function, and we assumed that \(F_i(\Phi)\leq \alpha\). This concludes the proof. 

\section{Proof of Proposition \ref{prop:FiEqualsZero}} %3.2
\label{sec:ProofFiEqualsZero}
We first show that if \(\Phi\) is a Markov potential function then \(F_i(\Phi)=0\) for every \(i\in \mathcal{N}\). By definition of potential game, for all players $i \in \mathcal{N}$,
\begin{align*}
    \Phi(\pi_i, \pi_{-i}) - \Phi(\pi'_i, \pi_{-i}) = J_i(\pi_i, \pi_{-i}) - J_i(\pi'_i, \pi_{-i}) \quad  \forall \pi'_i \in \Pi_i, \forall \pi_{-i} \in \Pi_{-i}.
\end{align*}
Selecting \(\pi_i' = \pi_i^{\ast,J}\) and \(\pi = \pi^{\ast,\Phi}\) in above equation shows that \(F_i(\Phi)= 0.\)

To prove there exists games that are not potential games but satisfy $F_i(\Phi) = 0$ for every $i \in \mathcal{N}$, we construct an example. The game in (\ref{eq:matrxigame}) is not a potential game (see Appendix \ref{sec:NotPotentialProof}). However, for $\Phi_\omega \in (0.6,1)$, $F_i(\Phi_\omega)=0$ for every \(i\in \mathcal{N}\), due to Proposition \ref{prop:toy_example} (proof in Appendix \ref{sec:toy_example_proof}). This shows an example exists, completing the proof.

\section{Proof of Proposition \ref{prop:toy_example}}
\label{sec:toy_example_proof}
We prove by constructing an example. From the definition of potential function \(\Phi_w\) in \eqref{eq:matrxigame}, we see 
\begin{align}
\label{eq:max_phi_ex}
    \max_{\pi}\Phi_w (\pi) = \begin{cases}
    2(1-w ) & \text{if} \ w \leq 1/3\\
        \frac{7-5w }{4} & \text{if} \ w \in [1/3,0.6] \\ 
        1 & \text{if} \ w \in [0.6,1]
    \end{cases}
\end{align}

Moreover, the maximizer of the potential function is given as follows: 
\begin{align*}
    \pi^{\ast,\Phi} = \begin{cases}
        (A_2,B_2) & \text{if} ~ w  < 1/3\\
        \{\alpha (A_2,B_1) + (1-\alpha)(A_2,B_2) : \alpha \in [0,1]\}& \text{if} ~ w  = 1/3\\
       (A_2,B_1)& \text{if} ~ w  \in (1/3,0.6)\\
        \{\beta(A_1,B_1)+(1-\beta)(A_2,B_1): \beta\in [0,1] \}& \text{if} ~ w  = 0.6 \\ 
        (A_1,B_1) & \text{if} ~ w  \in (0.6,1) \\ 
        \{\gamma(A_1,B_1)+(1-\gamma)(A_1,B_2): \gamma \in [0,1]\} & \text{if} ~ w  =1
    \end{cases} 
\end{align*}

Using \(\pi^{\ast,\Phi}\), we compute individual best responses as follows: 
\begin{align*}
    &\pi_1^{\ast,J} = \arg\max_{\pi_1\in \Delta([A_1, A_2])} J_1(\pi_1, \pi_2^{\ast,\Phi}) = A_1  \\
    &\pi_2^{\ast,J} = \arg\max_{\pi_2\in \Delta([B_1, B_2])} J_2(\pi_1^{\ast,\Phi}, \pi_2) = \begin{cases}
        B_2 & \text{if} ~ w  \leq 1/3\\
       B_2& \text{if} ~ w  \in (1/3,0.6)\\
        B_2 & \text{if} ~ w  = 0.6, \beta < 1/3 \\ 
        \{\delta B_1 + (1-\delta)B_2:\delta\in [0,1]\} & \text{if} ~ w  = 0.6, \beta = 1/3 \\ 
        B_1 & \text{if} ~ w  = 0.6, \beta > 1/3\\ 
        B_1 & \text{if} ~ w  \in (0.6,1]
    \end{cases} 
\end{align*}

Finally, we note that 
\begin{align*}
    &\Phi_{w }(\pi^{\ast,\Phi}) - \Phi_{w }(\pi_{1}^{\ast,J}, \pi^{\ast,\Phi}_2) \\&= \begin{cases}
        \Phi_w (A_2,B_2)-\Phi_w (A_1,B_2) & \text{if} ~ w  < 1/3\\
        \{\alpha (\Phi_w (A_2,B_1)-\Phi_w (A_1,B_1))~ + \\
        \quad(1-\alpha)(\Phi_{w }(A_2,B_2)-\Phi_{w }(A_1,B_2)) : \alpha \in [0,1]\}& \text{if} ~ w  = 1/3\\
       \Phi_{w }(A_2,B_1)-\Phi_{w }(A_1,B_1)& \text{if} ~ w  \in (1/3,0.6)\\
        \{(1-\beta)(\Phi_w (A_2,B_1)-\Phi_w (A_1,B_1)): \beta\in [0,1] \}& \text{if} ~ w  = 0.6 \\ 
       0 & \text{if} ~ w  \in (0.6,1) \\ 
        0 & \text{if} ~ w  =1
    \end{cases}
    \\&= \begin{cases}
       \frac{3-5w }{2} & \text{if} ~ w  < 1/3\\
        \{\alpha (\frac{3-5w }{4}) + (1-\alpha)(\frac{3-5w }{2}) : \alpha \in [0,1]\}& \text{if} ~ w  = 1/3\\
       \frac{3-5w }{4}& \text{if} ~ w  \in (1/3,0.6)\\
        \{(1-\beta)\frac{3-5w }{4}: \beta\in [0,1] \}& \text{if} ~ w  = 0.6 \\ 
       0 & \text{if} ~ w  \in (0.6,1) \\ 
        0 & \text{if} ~ w  =1
    \end{cases}. 
\end{align*}
Similarly, 
\begin{align*}
    &\Phi_{w }(\pi^{\ast,\Phi}) - \Phi_{w }(\pi_{1}^{\ast,\Phi}, \pi^{\ast,J}_2) \\&= \begin{cases}
        0 & \text{if} ~ w  < 1/3\\
        \{\alpha (\Phi_{w }(A_2,B_1)-\Phi_{w }(A_2,B_2)) : \alpha \in [0,1]\}& \text{if} ~ w  = 1/3\\
       (\Phi_{w }(A_2,B_1)-\Phi_{w }(A_2,B_2))& \text{if} ~ w  \in (1/3,0.6)\\
        \{\beta(\Phi_{w }(A_1,B_1)-\Phi_{w }(A_1,B_2))~+ \\
        \quad(1-\beta)(\Phi_{w }(A_2,B_1)-\Phi_{w }(A_2,B_2)): \beta\in [0,1] \}& \text{if} ~ w  = 0.6, \beta<1/3
        \\
        \{(1-\delta)\frac{1}{3}(\Phi_{w }(A_1,B_1)-\Phi_{w }(A_1,B_2))~+\\
        \quad \frac{2}{3}(\Phi_{w }(A_2,B_1)-\Phi_{w }(A_2,B_2))): \delta\in [0,1] \}& \text{if} ~ w  = 0.6, \beta=1/3
        \\
        0& \text{if} ~ w  = 0.6, \beta>1/3\\ 
        0 & \text{if} ~ w  \in (0.6,1) \\ 
        \{(1-\gamma)(\Phi_{w }(A_1,B_2)-\Phi_{w }(A_1,B_1)): \gamma \in [0,1]\} & \text{if} ~ w  =1
    \end{cases} 
    \\&= \begin{cases}
        0 & \text{if} ~ w  < 1/3\\
        0& \text{if} ~ w  = 1/3\\
       \frac{3w -1}{4}& \text{if} ~ w  \in (1/3,0.6)\\
       \frac{1}{5}& \text{if} ~ w  = 0.6, \beta<1/3
        \\
        \{(1-\delta)\left(\frac{1}{3}(\frac{1-w }{2})+\frac{2}{3}(\frac{3w -1}{4})\right): \delta\in [0,1] \}& \text{if} ~ w  = 0.6, \beta=1/3
        \\
        0& \text{if} ~ w  = 0.6, \beta>1/3\\ 
        0 & \text{if} ~ w  \in (0.6,1) \\ 
        0 & \text{if} ~ w  =1
    \end{cases}.
\end{align*}

Furthermore, 
\begin{align*}
    &J_1(\pi^{\ast,\Phi}) - J_1(\pi_{1}^{\ast,J}, \pi^{\ast,\Phi}_2)\\& = \begin{cases}
        J_1(A_2,B_2)-J_1(A_1,B_2) & \text{if} ~ w  < 1/3\\
        \{\alpha (J_1(A_2,B_1)-J_1(A_1,B_1))~+\\
        \quad(1-\alpha)(J_1(A_2,B_2)-J_1(A_1,B_1)) : \alpha \in [0,1]\}& \text{if} ~ w  = 1/3\\
       J_1(A_2,B_1)-J_1(A_1,B_1)& \text{if} ~ w  \in (1/3,0.6)\\
        \{(1-\beta)(J_1(A_2,B_1)-J_1(A_1,B_1)): \beta\in [0,1] \}& \text{if} ~ w  = 0.6 \\ 
        0 & \text{if} ~ w  \in (0.6,1) \\ 
        0 & \text{if} ~ w  =1
    \end{cases} 
    \\& = \begin{cases}
       -1 & \text{if} ~ w  < 1/3\\
        \{\alpha (-0.5) + (1-\alpha)(-1) : \alpha \in [0,1]\}& \text{if} ~ w  = 1/3\\
      -0.5& \text{if} ~ w  \in (1/3,0.6)\\
        \{(1-\beta)(-0.5): \beta\in [0,1] \}& \text{if} ~ w  = 0.6 \\ 
        0 & \text{if} ~ w  \in (0.6,1) \\ 
        0 & \text{if} ~ w  =1
    \end{cases},
\end{align*}
and
\begin{align*}
    &J_2(\pi^{\ast,\Phi}) - J_2(\pi_{1}^{\ast,\Phi}, \pi^{\ast,J}_2) \\&
    =\begin{cases}
        0 & \text{if} ~ w  < 1/3\\
        \{\alpha (J_2(A_2,B_1)-J_2(A_2,B_2)) : \alpha \in [0,1]\}& \text{if} ~ w  = 1/3\\
       (J_2(A_2,B_1)-J_2(A_2,B_2))& \text{if} ~ w  \in (1/3,0.6)\\
        \{\beta(J_2(A_1,B_1)-J_2(A_1,B_2))~+\\
        \quad(1-\beta)(J_2(A_2,B_1)-J_2(A_2,B_2)): \beta\in [0,1] \}& \text{if} ~ w  = 0.6, \beta<1/3
        \\
        \{(1-\delta)(\frac{1}{3}(J_2(A_1,B_1)-J_2(A_1,B_2))~+\\
        \quad\frac{2}{3}(J_2(A_2,B_1)-J_2(A_2,B_2))): \delta\in [0,1] \}& \text{if} ~ w  = 0.6, \beta=1/3
        \\
        0& \text{if} ~ w  = 0.6, \beta>1/3\\ 
        0 & \text{if} ~ w  \in (0.6,1) \\ 
        \{(1-\gamma)(J_2(A_1,B_2)-J_2(A_1,B_1)): \gamma \in [0,1]\} & \text{if} ~ w  =1
    \end{cases} \\&
    =\begin{cases}
        0 & \text{if} ~ w  < 1/3\\
        \{\alpha (-1/4) : \alpha \in [0,1]\}& \text{if} ~ w  = 1/3\\
       (-1/4)& \text{if} ~ w  \in (1/3,0.6)\\
        \{\beta(1/2)+(1-\beta)(-1/4): \beta\in [0,1] \}& \text{if} ~ w  = 0.6, \beta<1/3
        \\
        \{(1-\delta)\left(\frac{1}{3}(1/2)+\frac{2}{3}(-1/4)\right): \delta\in [0,1] \}& \text{if} ~ w  = 0.6, \beta=1/3
        \\
        0& \text{if} ~ w  = 0.6, \beta>1/3\\ 
        0 & \text{if} ~ w  \in (0.6,1) \\ 
        \{(1-\gamma)(-1/2): \gamma \in [0,1]\} & \text{if} ~ w  =1
    \end{cases} 
\end{align*}

Combining everything together, we obtain
\begin{align*}
    &F_1 = \Phi_{w }(\pi^{\ast,\Phi}) - \Phi_{w }(\pi_{1}^{\ast,J}, \pi^{\ast,\Phi}_2) - (J_1(\pi^{\ast,\Phi}) - J_1(\pi_{1}^{\ast,J}, \pi^{\ast,\Phi}_2) ) = \\
    &\begin{cases}
       \frac{5-5w }{2} & \text{if} ~ w  < 1/3\\
        \{\alpha (\frac{5-5w }{4}) + (1-\alpha)(\frac{5-5w }{2}) : \alpha \in [0,1]\}& \text{if} ~ w  = 1/3\\
       \frac{5-5w }{4}& \text{if} ~ w  \in (1/3,0.6)\\
        \{(1-\beta)\left(\frac{5-5w }{4}\right): \beta\in [0,1] \}& \text{if} ~ w  = 0.6 \\ 
       0 & \text{if} ~ w  \in (0.6,1) \\ 
        0 & \text{if} ~ w  =1
    \end{cases}
\end{align*}

{\begin{align*}
    &F_2 = \Phi_{w }(\pi^{\ast,\Phi}) - \Phi_{w }(\pi_{1}^{\ast,\Phi}, \pi^{\ast,J}_2) - (J_2(\pi^{\ast,\Phi}) - J_2(\pi_{1}^{\ast,\Phi}, \pi^{\ast,J}_2) ) = \\
    &=\begin{cases}
        0 & \text{if} ~ w  < 1/3\\
        0+\{\alpha (-1/4) : \alpha \in [0,1]\}& \text{if} ~ w  = 1/3\\
       \frac{3w }{4}& \text{if} ~ w  \in (1/3,0.6)\\
       \frac{1}{5}-\{\beta(1/2)+(1-\beta)(-1/4): \beta\in [0,1] \}& \text{if} ~ w  = 0.6, \beta<1/3
        \\
        \{(1-\delta)\left(\frac{1}{3}(\frac{-w }{2})+\frac{2}{3}(\frac{3w }{4})\right): \delta\in [0,1] \}& \text{if} ~ w  = 0.6, \beta=1/3
        \\
        0& \text{if} ~ w  = 0.6, \beta>1/3\\ 
        0 & \text{if} ~ w  \in (0.6,1) \\ 
        \{(1-\gamma)(-1/2): \gamma \in [0,1]\} & \text{if} ~ w  =1
    \end{cases}
\end{align*}}

\begin{align*}
    &\max\{F_1,F_2\}(w ) = 
    \begin{cases}
        5\frac{1-w }{2} & \text{if} ~ w  < 1/3\\
        \{\frac{5}{6}(2-\alpha) : \alpha \in [0,1]\}& \text{if} ~ w  = 1/3\\
      5\frac{1-w }{4} & \text{if} ~ w  \in (1/3,0.6)\\
        \{\frac{1-\beta}{2}:\beta\in[0,1]\}& \text{if} ~ w  = 0.6, 
        \\ 
        0 & \text{if} ~ w  \in (0.6,1] 
    \end{cases}
\end{align*}

\section{Proof that game in \eqref{eq:matrxigame} is not a potential game}
\label{sec:NotPotentialProof}
Here, we show that the game in \eqref{eq:matrxigame} is not a potential game. Let's prove this by contradiction. Suppose it were a potential game, then there exists  a function \(\Phi\) such that  
\begin{equation}\label{eq:poot1}
\begin{aligned}
    \Phi(A_1,B_1) - \Phi(A_2,B_1) = J_1(A_1,B_1)-J_1(A_2,B_1) = 1/2 \\
    \Phi(A_1,B_2) - \Phi(A_2,B_2) = J_1(A_1,B_2)-J_1(A_2,B_2) =  1
\end{aligned}
\end{equation}
and 
\begin{equation}\label{eq:poot2}
\begin{aligned}
    \Phi(A_1,B_1) - \Phi(A_1,B_2) = J_2(A_1,B_1)-J_2(A_1,B_2) = 1/2 \\
    \Phi(A_2,B_1) - \Phi(A_2,B_2) = J_2(A_2,B_1)-J_2(A_2,B_2) = -1/4 
\end{aligned}
\end{equation}

From \eqref{eq:poot1}, we have that  
\begin{align}\label{eq:fac1}
    \Phi(A_1,B_1) - \Phi(A_2,B_1) - \Phi(A_1,B_2) + \Phi(A_2,B_2) = -1/2.
\end{align}

Moreover, from \eqref{eq:poot2}, we have that  
\begin{align}\label{eq:fac2}
    \Phi(A_1,B_1) - \Phi(A_1,B_2) - \Phi(A_2,B_1) + \Phi(A_2,B_2) = 3/4
\end{align}

But \eqref{eq:fac1} and \eqref{eq:fac2} contradict. This shows that the game in \eqref{eq:matrxigame} is not a potential game. 

\end{document}